# Estimating oil and gas recovery factors via machine learning: Database-dependent accuracy and reliability

Alireza Roustazadeh[1], Behzad Ghanbarian[1*], Mohammad B. Shadmand[2], Vahid Taslimitehrani[3], and Larry W. Lake[4]

[1] Porous Media Research Lab, Department of Geology, Kansas State University, Manhattan 66506 KS, United States

[2] Department of Electrical and Computer Engineering, College of Engineering, University of Illinois at Chicago, Chicago 60607 IL, United States

[3] Staff Machine Learning Scientist at Realtor.com, San Francisco CA, United States

[4] Hildebrand Department of Petroleum and Geosystems Engineering, University of Texas at Austin, Austin 78712 TX, United States

* Corresponding author's email address: ghanbarian@ksu.edu

**Abstract**

With recent advances in artificial intelligence, machine learning (ML) approaches have become an attractive tool in petroleum engineering, particularly for reservoir characterizations. A key reservoir property is hydrocarbon recovery factor (RF) whose accurate estimation would provide decisive insights to drilling and production strategies. Therefore, this study aims to estimate the hydrocarbon RF for exploration from various reservoir characteristics, such as porosity, permeability, pressure, and water saturation via the ML. We applied three regression-based models including the extreme gradient boosting (XGBoost), support vector machine (SVM), and stepwise multiple linear regression (MLR) and various combinations of three databases to construct ML models and estimate the oil and/or gas RF. Using two databases and the cross-validation method, we evaluated the performance of the ML models. In each iteration 90 and 10% of the data were respectively used to train and test the models. The third independent database was then used to further assess the constructed models. For both oil and gas RFs, we found that the XGBoost model estimated the RF for the train and test datasets more accurately than the SVM and MLR models. However, the performance of all the models were unsatisfactory for the independent databases. Results demonstrated that the ML algorithms were highly dependent and sensitive to the databases based on which they were trained. Statistical tests revealed that such unsatisfactory performances were because the distributions of input features and target variables in the train datasets were significantly different from those in the independent databases (p-value < 0.05).
**Keywords:** Hydrocarbon, Machine learning, Recovery factor, Regression, XGBoost

**1. Introduction**

Successful exploration and production of a hydrocarbon reservoir have always been challenging in petroleum engineering [1]. Because of high fluctuations in oil and gas prices, it is necessary to determine, if a field is economically feasible to be invested or not. A conventional reservoir goes through different stages during its lifetime, namely exploration, appraisal, development, production, and lastly abandonment. Once a reservoir is discovered (exploration stage), the main challenge is to determine whether it has high potential to produce substantial amount of hydrocarbon and subsequently yield enough profit. Even for a currently producing reservoir, one still needs to analyze its performance and production rate to assure profits above economical margins.

The performance of a reservoir influences its economic feasibility and is a function of reservoir quality. Reservoir performance may be quantified by the initial production rate of a reservoir and the decline in its production rate. Another indicator is the amount of recovered hydrocarbon from the initial volume of hydrocarbon in a reservoir [2], also known as recovery factor (RF). The RF, ranging between 0 and 1, is a quantity to evaluate the fate of a reservoir. It is a function of displacement mechanisms, such as water drive, gas cap, rock compaction drive, solution gas, and gravity drainage [3]. Its value may be determined at different stages of a reservoir lifetime. However, in this study we refer to *ultimate* recovery factor by RF.

There exist different methods to determine the RF during the lifetime of a reservoir. When a field is in its appraisal phase and no production data are available, analogs and empirical formulas, e.g., history matching and volumetric reserve estimation, are commonly used. Such methods, however, generally come with substantial uncertainties due to the lack of adequate analogs [4]. During the production phase, dynamic properties of a field vary with production and time and, thus, the RF determination is influenced by enhanced and improved recovery methods [5] as well as more data being collected from a field [6]. Once approaching the abandonment phase, the emphasis should be more on economical margin, and how close the field production is to such a margin. At the plateau production phase, the RF is typically determined by simulations and dynamic reservoir modeling. However, at late stages of production and when the plateau production is no longer in place, the decline curve analysis should be analyzed. Various methods used at different field life stages may lead to different RF determinations and uncertainties [4]. In addition to the decline curve analysis, one may apply the material balance method to determine the reserve and, accordingly, the recovery factor [4,6,7]. Both the material balance and production decline curve analysis methods calculate reserves and recovery factor from the performance of a field [8,9].

The aforementioned methods are time consuming, and their results may have a wide range of uncertainties[10–12]. Moreover, the RF is a function of multiple variables e.g., permeability, temperature, and gas oil ratio (GOR), which may not be necessarily incorporated in such methods. In addition, those methods are not cost efficient. With recent advances in artificial intelligence and data analytics, one may construct a machine learning-based model to estimate the RF from other available reservoir characteristics [13]. Such models may provide a cost efficient and accurate platform as well as better insights into reservoir characterization and hydrocarbon production, if implemented appropriately.

Machine learning (ML) techniques have been previously used to estimate the RF [14–17]. For a recent review see Tahmasebi et al. [18]. For example, Lee and Lake [19] applied various methods including multiple linear regression (MLR), MLR with sequential feature selection, artificial neural networks (ANN), and Bayesian network to estimate both oil and gas RFs. In their study, the ANN and MLR with sequential feature selection showed better performance than the other two methods. Aliyuda and Howell [4] applied the MLR and support vector machine (SVM) with Gaussian kernel on 93 reservoirs from the Norwegian Continental Shelf. 75 reservoirs were from the Norwegian Sea, the Norwegian North Sea, and the Barents Sea, and the remaining 18 reservoirs from the Viking Graben in the UK sector of the North Sea. Aliyuda and Howell [4] found that the SVM model outperformed the MLR model.

In another study, Chen et al. [16] applied the ANN approach to develop predictive oil RF models with different sets of input data using the database TORIS and identified 19 principal features (out of 70) in the construction of their ML model. Tewari et al. [20] used six different approaches, i.e., random tree, random forest, SVM, bagging, radial basis function, and multilayer perceptron and compared them with their proposed ensemble estimator (E2) model. Ensemble methods are classifier algorithms that develop multiple classifiers and make inferencing using the weighted vote of multiple estimations that they have made [21]. Tewari et al. [20] showed that their E2-based model outperformed the other six methods.

In the literature, most ML studies divide a database into train and test splits. Although the performance of ML models is evaluated using the test dataset, an independent database with no data overlap is not used to further assess the accuracy and reliability of ML models constructed. Furthermore, to the best of the authors' knowledge, the extreme gradient boosting algorithm has neither been applied to estimate the hydrocarbon recovery factory at the reservoir scale, nor has it been compared with support vector machine and multiple linear regression. Therefore, the main objectives of this study are to: (1) apply the extreme gradient boosting (XGBoost) approach to develop an ML-based model, (2) compare its performance with that of multiple linear regression (MLR) and support vector machine (SVM) in the estimation of oil and gas RFs using reservoirs from around the world, and (3) address the database dependence of accuracy and uncertainty in ML-based models.

## 2. Materials and Methods
### 2.1. Databases
**- Commercial database**

The commercial database analyzed in this study is the same one that [19] used. This database contains more than 1200 samples including conventional reservoirs with different geologic formations from around the world. The oil RF values were reported for more than 600 samples, while the gas RF values only available for more than 300 cases. For each sample representing a reservoir, there exist more than 200 numerical and categorical features, such as porosity, permeability, reservoir fractures, and recovery factor.

**- Bureau of Ocean Energy Management (BOEM) Atlas of Gulf of Mexico**

This database contains the conventional oil and gas reserves of the outer continental shelf region of Gulf of Mexico [22]. There exist more than 13000 samples and more than 80 features for each sample. This is the most complete database with only a few missing values. It includes samples with both oil and gas recovery factor values.

**- Tertiary Oil Recovery Information System (TORIS) database**

TORIS is a suite of databases gathered by the National Petroleum Council and used by the US Department of Energy to identify and characterize US oil conventional resources and further strategize their plans. It consists of more than 1300 reservoirs (samples) and more than 60 categorical and numerical features for each sample. TORIS is an old database strictly compiled for oil reservoirs and was used in combination with other databases to estimate the oil RF only.

**- Gas Information System (GasIS) database**

GasIS is a merged database, a combination of multiple pre-existing datasets. It is the first national public database containing information on multiple reservoir properties and ultimate reservoir recovery for conventional gas reservoirs. Created by the Department of Energy [23], GasIS consists of gas reservoirs with more than 19000 samples and 186 categorical and numerical features. The database, however, has many missing values. After cleaning up by omitting the samples with missing RF values, only 8524 samples were remained.

The Commercial and Atlas databases contain both oil and gas RF data, while the GasIS and TORIS databases only the gas and oil RF data, respectively. To avoid data overlap among the databases used in this study, we assured samples were presented in only one database and not repeated in other databases. Figure 1 presents the oil and gas RFs versus the permeability (millidarcy) and porosity (ft$^3$/ft$^3$) for the databases used in this study. This figure shows the range of each variable and the trend between the RF and permeability as well as porosity within each database. As can be seen, the plots are scattered, and there is no strong correlation between the RF and permeability or porosity.

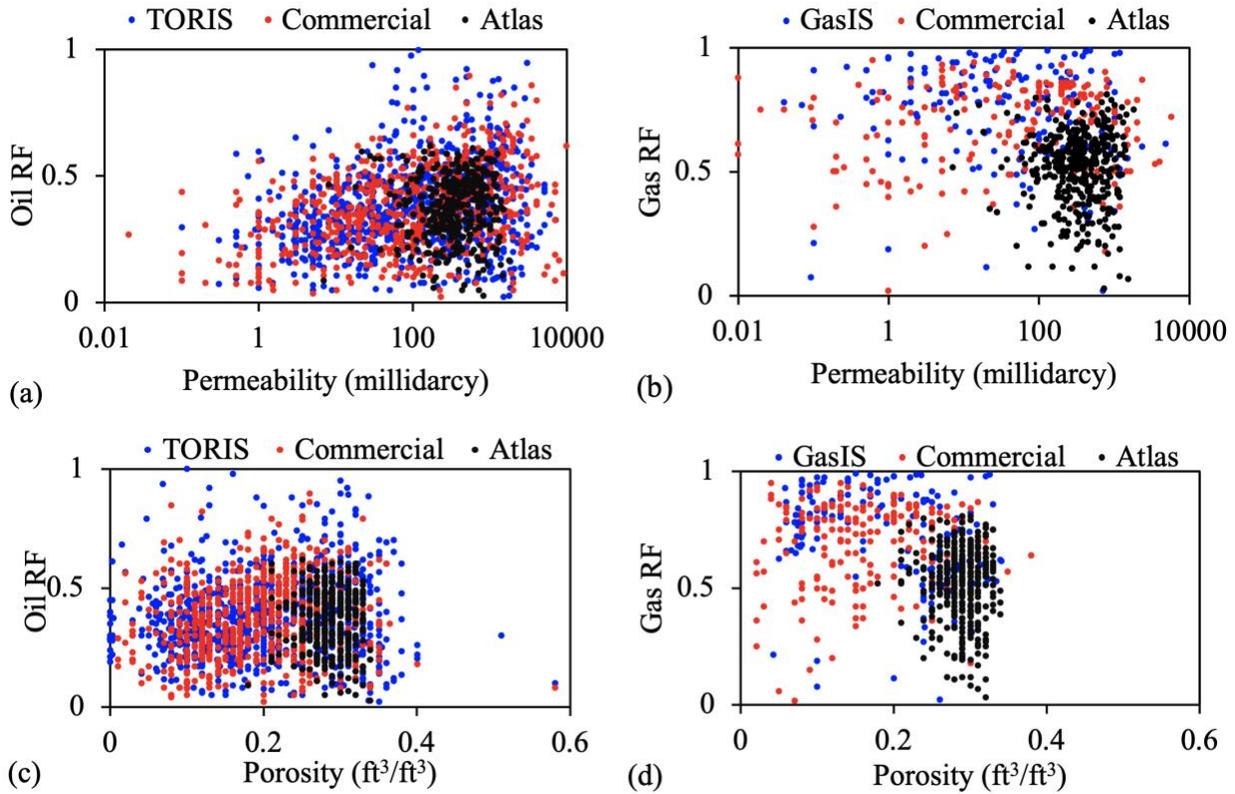

Figure 1. Oil and gas recovery factors against permeability (millidarcy) and porosity (ft$^3$/ft$^3$) for different databases used in this study.

To construct the ML-based models and address their database-dependent accuracy and reliability in the estimation of the oil and gas RFs, the databases collected in this study were merged using different combinations. For example, for the oil RF we combined the TORIS and Commercial databases (named TC), TORIS and Atlas databases (named TA), Commercial and Atlas databases (named CA), and all the three databases (named TCA). Similar approach was used for the gas RF; CA represents Commercial combined with Atlas, GC denotes GasIS combined with Commercial, GA is GasIS combined with Atlas, and GCA represents GasIS combined with commercial and Atlas. Whenever two databases were merged, the third one was used as an independent database for further evaluation of the constructed ML models.

**2.2. Data preparation**

**2.2.1. Feature engineering**

For the feature engineering, we first selected those features that were common in all the oil or gas databases e.g., porosity, permeability, pressure, reservoir thickness, temperature, and reserves. All the features that had reservoir information after the exploration phase e.g., production rate and cumulative production were removed simply because we aimed to propose a model estimating the RF from data collected at early stages of a reservoir's life. To ensure that the selected features were not *highly* correlated with each other, Spearman's rank correlation coefficient (also known as Spearman's Rho test) was used to determine the correlation between any two features. We should point out that collinearity is not of concern when decision trees are used [24]. Table 1 lists those features used in the ML algorithms and their limits in both oil and gas databases. Reserves in the GasIS and Atlas databases indicate the amount of hydrocarbon remaining in reservoirs that are commercially recoverable. Although it is not clear what reserves represent in the Commercial and TORIS databases, we assumed the same definition as in GasIS and Atlas.

Table 1. Input features and target variables selected to construct the machine learning-based models and numerical intervals used in oil and gas databases. Note that RF is the target variable, and the rest are input features.

| Oil | Oil databases | Gas | Gas databases |
|---|---|---|---|
| API Gravity<br>$B_o$ (RB/STB)<br>GOR* (MSCF/RB)<br>Water saturation<br>Temperature (°F)<br>Pressure (psi)<br>Thickness (ft)<br>Reserves (STB)<br>Permeability (mD)<br>Porosity (ft³/ft³)<br>Area (acre)<br>Oil RF | $1 \leq B_o \leq 3$<br>$0 \leq GOR \leq 60$<br>$0 \leq S_w \leq 1$<br>$0 \leq Reserves \leq 5\times 10^{11}$<br>$0 \leq \phi \leq 1$<br>$0 \leq Oil\ RF \leq 1$ | GOR (MSCF/RB)<br>Water saturation<br>Temperature (F)<br>Pressure (psi)<br>Thickness (ft)<br>Reserves (MMSCF)<br>Permeability (mD)<br>Porosity (ft³/ft³)<br>Area (acre)<br>Gas RF | $0 \leq GOR \leq 10^9$<br>$0 \leq S_w \leq 1$<br>$0 \leq Reserves \leq 4\times 10^8$<br>$0 \leq \phi \leq 1$<br>$0 \leq Gas\ RF \leq 1$ |

API: American Petroleum Institute, $B_o$: oil formation volume factor (which depends on temperature and pressure), GOR: Gas Oil Ratio, MSCF/RB: *Thousands* Standard Cubic feet per Reservoir Barrel, STB: Stock Tank Barrel, $S_w$: water saturation, $\phi$: porosity, MMSCF: *Million* Standard Cubic Feet, RF: recovery factor.

### 2.2.2. Train – test split

The next step after cleaning data and feature engineering is to split the data into the train and test datasets. To construct the ML models, the data were randomly divided into two splits; 90% to train the model and 10% to test it. The 90%-10% split was selected so that the ML algorithms have as many samples as possible for training. Jaiswal et al. [25] also used the same train and test split ratio in their study. For cross validation and optimizing the model parameters, the train dataset was used to create smaller datasets (called folds). All the preprocessing steps carried out on the test datasets and the independent databases were performed using the same parameters used for the train datasets [26]. This means that the imputation performed on the train and test datasets were performed separately using the mode of each feature specifically in each database. However, no imputation was applied to the independent databases. Instead, any sample with missing value was removed. This is because the goal of further evaluating the proposed ML models using the independent databases was to critically assess them based on real data not imputed ones. As stated earlier, samples within the independent databases were not available in either the train or test datasets (repeated samples were removed to avoid data overlap).

### 2.2.3. Missing value imputation

After selecting the features and splitting the data into the train and test datasets, we sorted the data based on the target feature (i.e., RF) and omitted samples with missing RF values followed by removing outliers to prepare the datasets for imputation. To do that, each feature was capped from a lower to an upper limit depending on each feature nature and its histogram. Finally, we removed any samples that had more than 55% of its values missing.

Missing value imputation is a common strategy employed to tackle incomplete datasets [27]. An oversimplified imputation method is to replace missing values of a feature with its mean, which typically results in rough approximations. In this study, however, we first sorted all the samples based on their RF values ascendingly. Then, we divided data for each feature into smaller subsamples (SS) of 10 entries and assured that the ratio (R) of the missing values (MS) to the total size of the sub-sample (TSS) was less than or equal to 10%. If not, more entries were added to meet the criterion (MS/TSS $\leq$ 0.1). We next replaced the missing values with the value with the highest frequency (mode) in that subsample. The logic behind this imputation method is to better estimate the missing values for one feature based on similarities in the RF values. The workflow of the proposed imputation method is summarized in Figure 2.

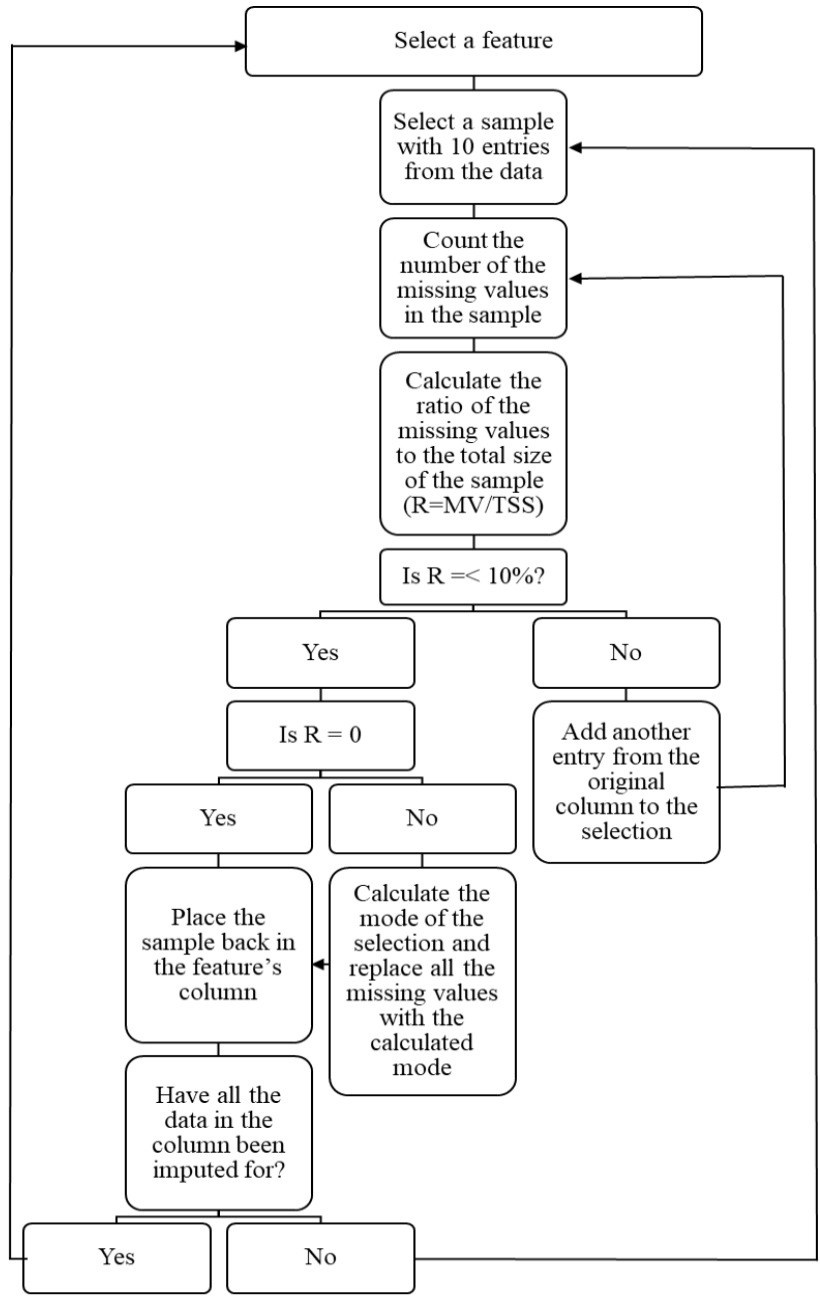

Figure 2. Imputation steps used to replace missing values in all the databases used to construct machine learning models.

## 2.2.4. Data standardization and normalization

To standardize the, we used the Gaussian rank transformation to convert the distribution of each feature to approximately Gaussian. This method ranks each value of a feature based on its magnitude from lowest to highest (larger values of higher ranks). To normalize the data and remove bias and minimize uncertainties, one should have dimensionless input and output features with similar ranges and magnitude variations between 0 and 1 [28]. For this purpose, using the minimum and maximum values of each feature, the input and output data were scaled using the equation $X_{norm} = (x-x_{min})/(x_{max}-x_{min})$ followed by $X_{scaled} = X_{norm} \times (max - min) + min$ [29] in which min and max are, respectively, desired lower and upper limit values. In this study, the limits were chosen to be 0 (min) and 1 (max).

## 2.3. Learning curve construction

In the ML, over- and/or under-fitting may happen because of insufficient or excessive number of training samples [30]. Overfitting means an ML model returns accurate estimations for train dataset, while fails to precisely determine target variable for test dataset. Underfitting means a constructed model cannot encompass and handle the intrinsic complexity of training and testing data [31]. To avoid it, we investigated learning curves for each algorithm by constructing a basic model using the first sample in the train dataset. Predictions were then carried out for the same sample in the train dataset as well as the first sample in the test dataset. Next, a similar basic model was constructed based on the first two samples in the train dataset and used to estimates the RF values for the first two samples in the train and test datasets. This process was iterated until all the samples in the train dataset were included in the basic model. The effect of adding samples was quantified by calculating the RMSE value. Figure 3 shows different schematic learning curves and over- or under-fitting interpretations. The closer the train and test learning curves are, the less the over-fitting (minimal to no over-fitting trend shown in black) or under-fitting (minimal to no over-fitting trend shown in green) would be. Note that the test learning curve far below and above the train learning curve, respectively, indicate substantial under- and over-fitting, shown by blue and brown colors on Fig. 3.

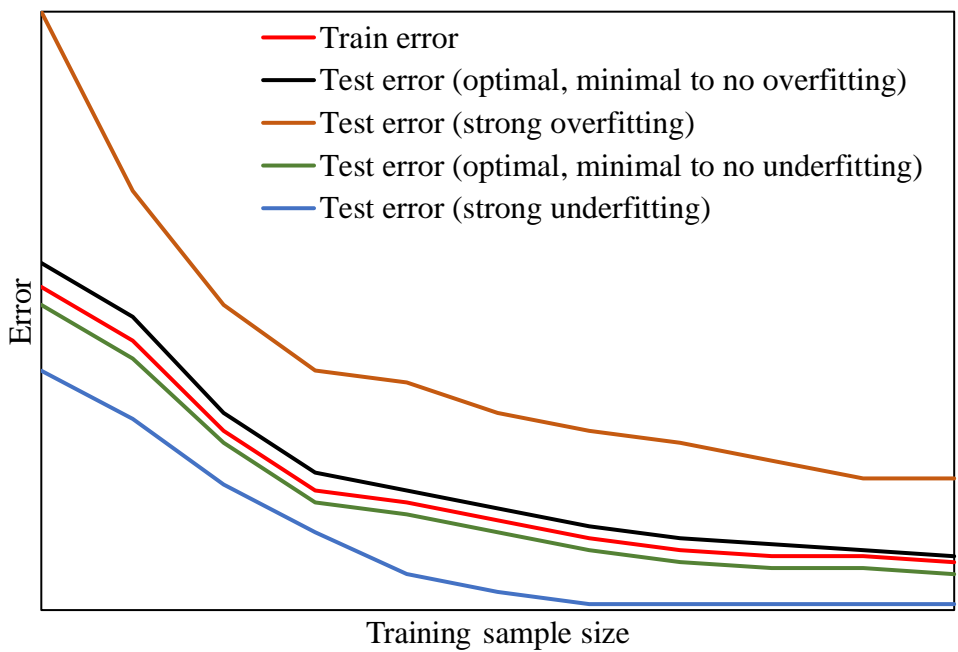

Figure 2. The scheme of different possible learning curves and their over/underfitting interpretations due to sample size. The test learning curve (shown in brown) far above the train learning curve (shown in red) indicates strong overfitting, while the test curve (shown in blue) far below the train curve denotes strong underfitting.

**2.4. Machine learning-based model construction**

In this study, we used the Python programming and applied extreme gradient boosting (XGBoost), support vector machine (SVM), and multiple linear regression (MLR) each of which is briefly described in the following. Although there are various ML algorithms, we selected these three because the XGBoost is among the latest ML algorithms developed. It is open access and has a user-friendly platform. The SVM is widely used in the petroleum industry for ML purposes [32–36]. The MLR is simple and among the basic ML models. It is to compare with the other two more complicated ML algorithms.

- **XGBoost**

The XGBoost is a decision tree based ensemble machine learning algorithm based on a gradient boosting [37]. It is a supervised algorithm and can be used for both regression and classification tasks [38,39]. The XGBoost is based on a gradient boosting framework that convert weak learners to strong ones in a sequential manner [40]. In each iteration, gradient boosting machines (GBM) identify weak learners and try to empower them using higher weights and eventually convert them to strong learners, which improves the performance. Compared to the GBM, the XGBoost is an improvised version that yields more accurate results with less amount of computing resources. Some of those optimizations are parallelized implementation and L1 and L2 regularization to avoid overfitting [37].

- **Support Vector Machine**

The SVM is a supervised algorithm that aims to construct a model to predict the category of unseen data using many different classes found in a set of training samples based on the generalized portrait algorithm [41,42]. In the support vector regression [43], hyperplanes would be constructed, which acts as decision boundaries and make estimations of an output. Such hyperplanes are created through kernels, functions that convert input information into desired type for a support vector hyperplane. Hyperplanes are constrained by boundary lines that are epsilon (ε) units away around the hyperplane to construct a border among samples. The SVM regression is carried out by finding the best fit hyperplane that houses most number of data and are within the ε value rather than minimizing the error between measured and estimated values [44].

- **Forward selection stepwise multiple linear regression**

The stepwise MLR is a simple and basic machine learning model in which those input variables significantly contributing to outputs are detected via the t-test and then used to establish a linear-based relationship [45]. In each iteration, a feature is added to the model until there is no statistically significant parameter left (forward stepwise). The performance of the model is optimized through this process of feature addition [46].

## 2.5. Hyperparameter tuning

In ML, hyperparameter tuning is a tool used to create the best model possible [47]. For the XGBoost, the hyperparameters tuning was implemented using an in-house code. At each step, two hyperparameters were optimized and cross validated using 10 folds. For the database TC, however, we used 3 folds due to the limited number of samples in this database. The hyperparameters tuning process was iterated until all the parameters were tuned.

For the SVM, the hyperparameters were optimized through a grid search using the 10-fold cross-validation pipeline except for the database TC for which we used 3 folds. Tables 2 and 3 present the name and value of the optimized parameters, respectively, for the XGBoost and SVM models and the CA database.

Table 2. Optimized hyperparameters of the XGBoost models developed for the oil and gas RFs obtained on the CA (Commercial combined with Atlas) database.

| Parameters | Oil RF | Gas RF |
|---|---|---|
| Max depth | 4 | 4 |
| Minimum child weight | 6 | 3 |
| Learning rate | 0.05 | 0.1 |
| Subsample | 0.9 | 0.9 |
| Column sample by tree | 1 | 1 |
| Objective | reg:squarederror | reg:squarederror |
| Evaluation metric | rmse | rmse |
| Alpha | 0.3 | 0.8 |
| Lambda | 0.04 | 0.08 |
| Column sample by level | 1 | 0.9 |
| Gamma | 0.01 | 0.01 |
| Max delta step | 0.1 | 0.1 |
| Number of boosting rounds | 999 | 999 |

Table 3. Optimized hyperparameters of the SVM models developed for the oil and gas RFs obtained on the CA (Commercial combined with Atlas) database.

| Parameters | Oil RF | Gas RF |
|---|---|---|
| C | 10 | 100 |
| Epsilon | 0.1 | 0.01 |
| Gamma | 1 | 1 |
| Kernel | rbf | rbf |

## 2.6. Model evaluation

To evaluate the accuracy and reliability of the developed ML-based models, the root mean square error (RMSE), Pearson's correlation coefficient ($r$), and coefficient of determination (CD) were calculated as follows

$$RMSE = \sqrt{\frac{\sum_{i=1}^{N}(x_e - x_m)^2}{N}} \quad (1)$$

$$r = \frac{Cov(x_e, x_m)}{\sigma_{xm}\sigma_{xe}} \quad (2)$$

$$CD = 1 - \frac{SS_{res}}{SS_{tot}} \quad (3)$$

where $x_e$ is the estimated value, $x_m$ is the measured value, $N$ is the number of samples, $Cov(x_m, x_e)$ is the covariance of variables $x_m$ and $x_e$, $\sigma_{xm}$ and $\sigma_{xe}$ are the standard deviations of measured and estimated values, respectively, $SS_{res}$ is the residual sum of squares, and $SS_{tot}$ is the total sum of squares. One should note that based on the definition of CD given in Eq. (3), its value could be maximum 1 or negative depending on the ratio $SS_{res}/SS_{tot}$ [48].

## 2.7. Feature importance

The feature importance was carried out using the SHapley Additive exPlanations (SHAP) for the XGBoost and SVM models. SHAP describes how input features contribute to the estimation of a target variable [49]. Based on the theory proposed by Shapley [50], SHAP value determines the output of a game according to each player's contribution. Positive SHAP values indicate positive impacts on model estimations, while negative SHAP values denote negative impacts.

The feature importance for the MLR models was performed by investigating the regression coefficient of the input features [51]. This is because both the input and output variables were scaled between 0 and 1, and, thus, the greater the coefficient would be, the more important the feature is.

**2.8. Workflow of the constructed models**

Figure 4 summarizes the workflow including the steps explained above. This includes data preparation (section 2.2), learning curve construction (section 2.3), machine learning models (section 2.4), hyperparameter tuning (section 2.5), model evaluation (section 2.6), and feature importance (section 2.7). All models were constructed using the same procedure but different ML algorithms (i.e., XGBoost, SVM, and MLR) and combinations of databases. After each model was constructed, it was used to estimate the oil and/or gas RFs for the train and test datasets as well as the independent database. Overall, 12 models for oil and 12 models for gas RF estimation were developed using the three algorithms (24 models in total). Each model was evaluated statistically using the RMSE, CD, and r values and visually using the scatterplots and the spread of data points around the 1:1 line.

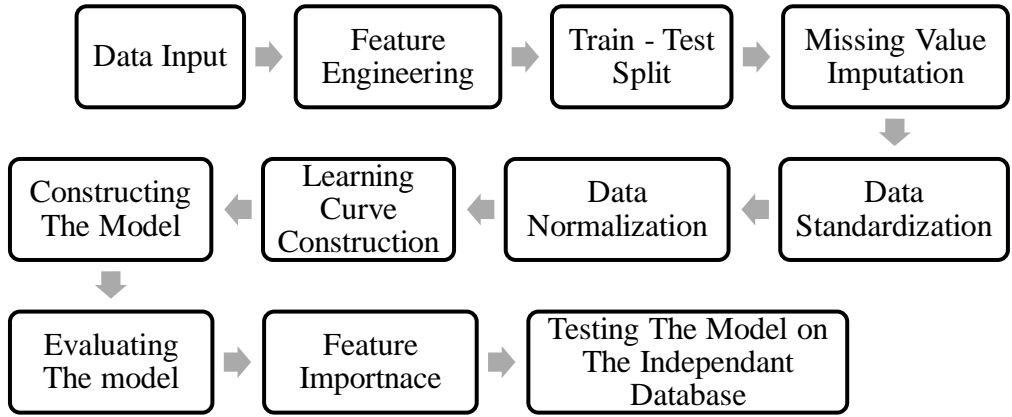

Figure 4. The workflow used here to develop the ML-based models and evaluate them. Standardization means converting the distribution of a feature to become approximately Gaussian, while normalization means scaling data between 0 and 1.

## 3. Results

### 3.1. Learning curves

Figure 5 shows the learning curves for the XGBoost models developed for estimating the oil and gas RFs using the CA database. At lower training sample sizes (< 3000), the test learning curves are far above the training learning curves indicating over-fitting for both oil and gas RFs. However, at greater training sample sizes, both training and test learning curves converge to some constant value close to each other. Results presented in Figure 5 indicate that the size of the train dataset was large enough. Furthermore, the RMSE value of the train dataset near the RMSE value of the test dataset demonstrates that the constructed models were not over- or under-fitted when the entire train dataset was used. Similar results were obtained for the other algorithms and databases (results not shown).

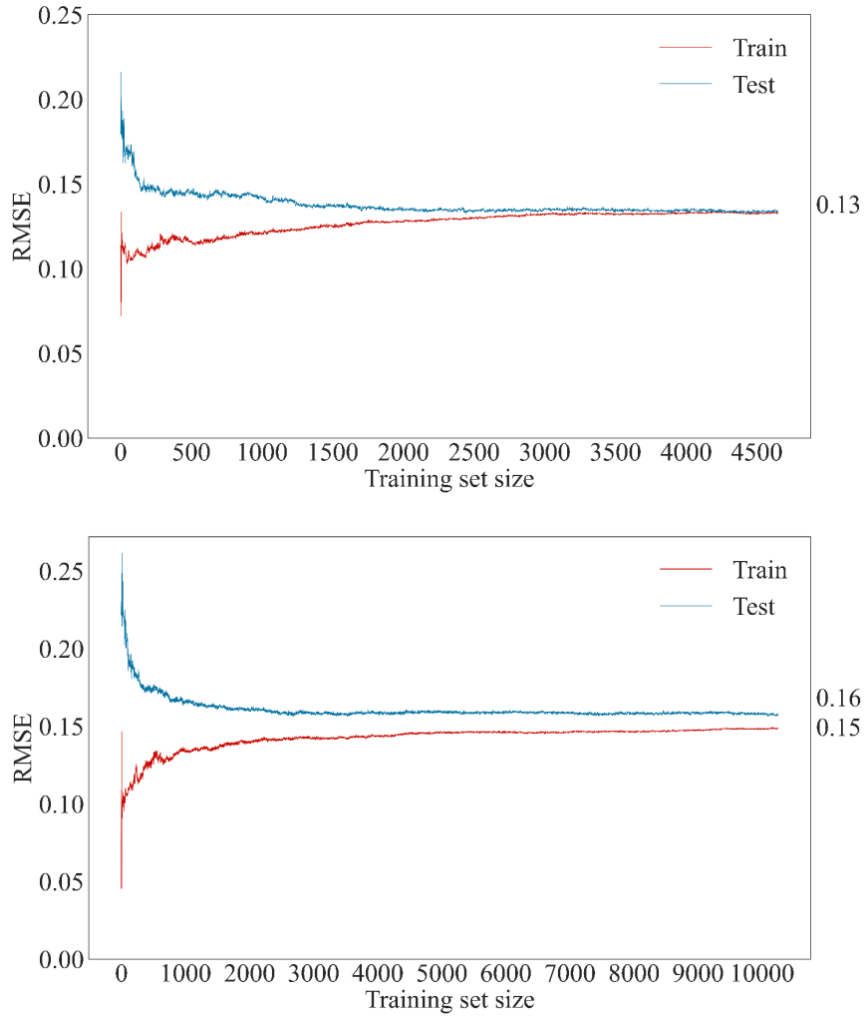

Figure 5. Learning curves of the XGBoost models developed for the (top) oil and (bottom) gas RFs using the CA (Commercial combined with Atlas) database.

**3.2. Oil RF estimations**

The RMSE, CD, and r values for the XGBoost, SVM, and MLR models and different combinations of the oil RF databases are summarized in Table 4. Comparing the RMSE, CD, and r values reported for the train datasets and different models shows that the XGBoost approach estimated the oil RF more accurately than the SVM and MLR models. Such values for the test datasets indicate that both the XGBoost and SVM models estimated the oil RF with similar accuracy but better than the MLR model. Results including the RMSE, CD, and r values for the independent databases, however, demonstrate that the accuracy of all the three models was almost similar (Table 4). Recall that the TCA database contains all the data from the TORIS, Commercial, and Atlas databases. Therefore, the independent database results are not available for this combination. The negative CD values obtained for the independent databases indicate that the constructed models could not accurately estimate the RF for new data from a different database. CD < 0 means that the estimations are worse than those based on the average value of the measured data. From Eq. (3), one should expect CD < 0 when the $SS_{res}$ value is greater than the $SS_{tot}$ value. The small Pearson's correlation coefficient values (r < 0.30) also indicate that the estimated RFs were weakly correlated to the measured values in the independent databases.

Figure 6 shows the results of the oil RF estimated via the XGBoost method against the measured value for the train and test datasets as well as the independent databases. As can be seen, the estimations are around the 1:1 line for the train and test datasets. Generally speaking, the value of RMSE decreases from top to bottom; RMSE = 0.120 (train) and 0.134 (test) for the TC database, while 0.035 (train) and 0.111 (test) for the TCA database. The TCA database is the combination of three databases with the largest number of samples. Results imply that the performance of the XGBoost model increased as the number of samples in the train dataset increased. Figure 6 also shows the results for the independent databases and the combinations TC, TA, and CA. Substantial greater RMSE values reported for the independent databases compared to those for the train and test datasets presented in Table 4 and Figure 6 clearly demonstrate that the ML models depend on the databases used to construct them. This means although all the models provided reasonable estimates of the oil RF for the test datasets, they failed to accurately estimate the RF for the independent databases. We address the database-dependent accuracy and reliability of the proposed models in further detail in the Discussion section.

Table 4. Calculated RMSE, CD and r values for the three algorithms XGBoost, SVM, and MLR used to estimate the oil RF. The train datasets of TC, TA, CA, and TCA databases consisted of 1502, 4779, 4654, and 5461 number of samples, respectively.

| Algorithm | XGBoost Regressor | | | | | | | | |
|---|---|---|---|---|---|---|---|---|---|
| Dataset | Train | | | Test | | | Independent | | |
| Database | RMSE | CD | r | RMSE | CD | r | RMSE | CD | r |
| TC | 0.120 | 0.45 | 0.70 | 0.134 | 0.31 | 0.57 | 0.174 | -0.14 | 0.17 |
| TA | 0.071 | 0.81 | 0.91 | 0.112 | 0.53 | 0.73 | 0.204 | -0.65 | 0.18 |
| CA | 0.065 | 0.84 | 0.93 | 0.100 | 0.61 | 0.78 | 0.260 | -1.50 | -0.04 |
| TCA | 0.035 | 0.96 | 0.98 | 0.111 | 0.56 | 0.75 | - | - | - |
| Algorithm | Support Vector Machine Regressor | | | | | | | | |
| Dataset | Train | | | Test | | | Independent | | |
| Database | RMSE | CD | r | RMSE | CD | r | RMSE | CD | r |
| TC | 0.135 | 0.30 | 0.55 | 0.135 | 0.29 | 0.55 | 0.169 | -0.08 | 0.29 |
| TA | 0.101 | 0.63 | 0.80 | 0.113 | 0.53 | 0.74 | 0.256 | -1.62 | 0.09 |
| CA | 0.100 | 0.61 | 0.78 | 0.100 | 0.61 | 0.78 | 0.285 | -1.97 | 0.02 |
| TCA | 0.110 | 0.57 | 0.75 | 0.130 | 0.41 | 0.65 | - | - | - |
| Algorithm | Stepwise Multiple Linear Regression | | | | | | | | |
| Dataset | Train | | | Test | | | Independent | | |
| Database | RMSE | CD | r | RMSE | CD | r | RMSE | CD | r |
| TC | 0.146 | 0.19 | 0.43 | 0.148 | 0.15 | 0.40 | 0.169 | -0.08 | 0.30 |
| TA | 0.125 | 0.46 | 0.65 | 0.122 | 0.42 | 0.66 | 0.205 | -0.68 | 0.19 |
| CA | 0.123 | 0.44 | 0.66 | 0.117 | 0.46 | 0.68 | 0.261 | -1.50 | 0.02 |
| TCA | 0.129 | 0.38 | 0.62 | 0.134 | 0.37 | 0.61 | - | - | - |

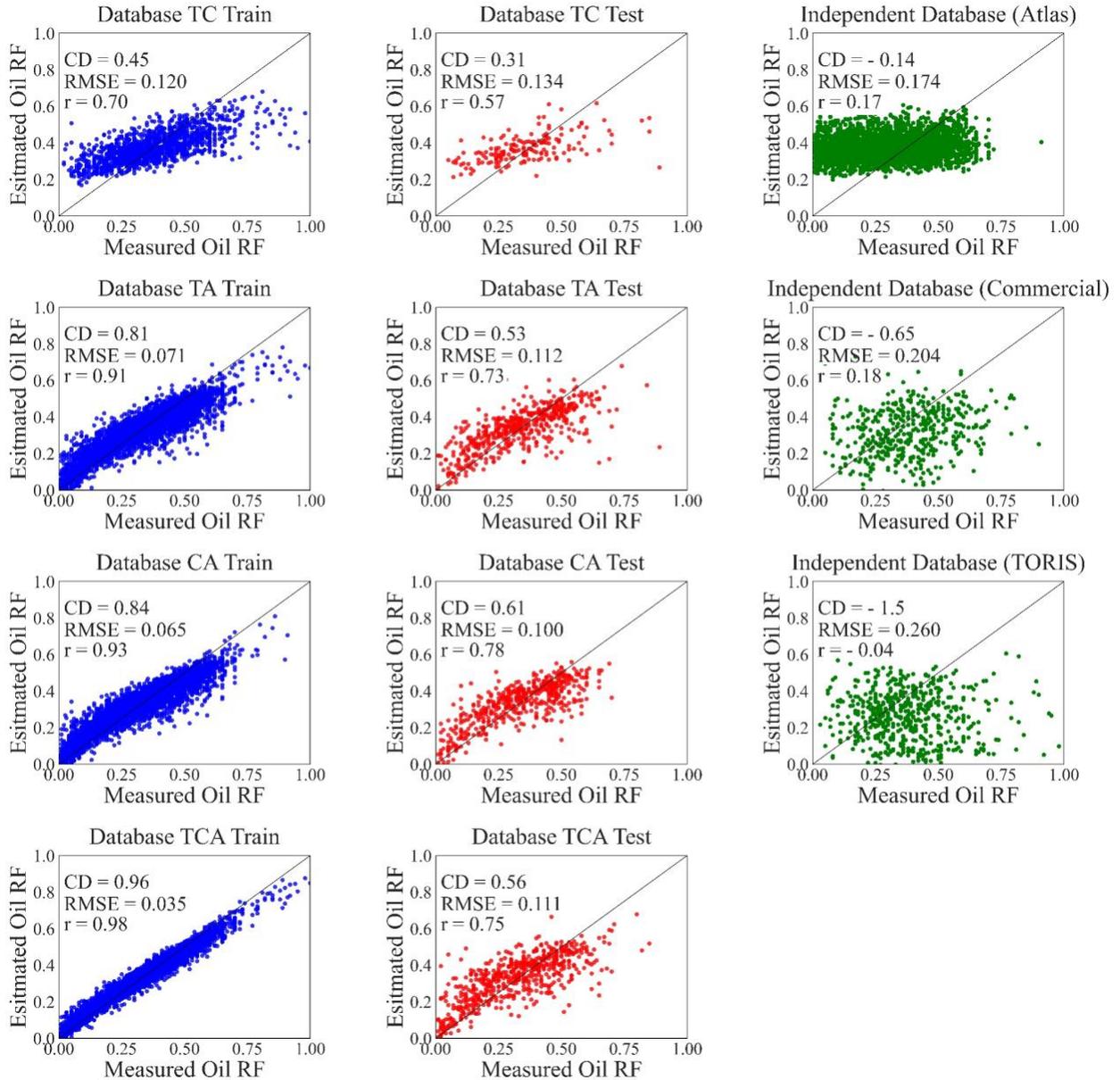

Figure 6. The estimated oil RF by the XGBoost algorithm versus that measured for the train and test datasets as well as the independent database. CA represents Commercial combined with Atlas, TC denotes TORIS combined with Commercial, TA is TORIS combined with Atlas, and CTA represents Commercial combined with TORIS and Atlas.

For the combination TC shown in Figure 6, the independent database is Atlas for which the estimated oil RFs are mainly restricted between 0.2 and 0.6, while the measured values mostly vary from 0 to 0.75. The reason for such a trend in the oil RF estimations is not clear yet. This, however, could be because the constructed model using the TC database estimates the oil RF for the train and test datasets mainly in the range of 0.2 and 0.6 (Figure 6).

Generally speaking, the XGBoost algorithm tends to overestimate at the lower oil RFs, while underestimate at the higher RF values for the train and test datasets, as depicted in Figure 6. Similar trend were reported by Talluru and Wu [13], and Lee and Lake [42] using other ML algorithms. Such a trend may be attributed to the fact that the databases used in this study contained samples with similar oil RFs, while their input features were substantially different. Such a complexity in data would decrease the efficiency of ML algorithms to adequately find and learn patterns among input features and target variables. Table 5 lists two samples with different oil RFs but similar input values and two other samples with similar oil RFs but different input features.

Table 5. Two samples with similar input features but different oil RFs (top two) and two samples with different input features but similar oil RFs (bottom two). Note that the reported values for the input features are not normalized (not scaled between 0 and 1) here.

| API gravity | $B_o$ (RB/STB) | GOR (MSCF/RB) | $S_w$ (-) | T (°F) | P (psi) | Thickness (ft) | Reserves (STB) | k (mD) | $\phi$ (ft³/ft³) | Area (acre) | Oil RF |
|---|---|---|---|---|---|---|---|---|---|---|---|
| 35 | 1.5 | 0.75 | 0.56 | 212 | 3823 | 141.1 | $2.46 \times 10^9$ | 200 | 0.23 | 20501 | 0.59 |
| 37 | 1.1 | 0.7 | 0.25 | 226 | 4357 | 141 | $2.56 \times 10^9$ | 25 | 0.14 | 20501 | 0.15 |
| 30 | 1.3 | 0.54 | 0.33 | 171 | 2404 | 673 | $1.54 \times 10^{10}$ | 1 | 0.09 | 73885 | 0.3 |
| 42 | 2 | 0.95 | 0.35 | 281 | 5163 | 10 | $2.4 \times 10^6$ | 62 | 0.16 | 742 | 0.3 |

API: American Petroleum Institute, $B_o$: oil formation volume factor, GOR: gas oil ratio, $S_w$: water saturation, T: temperature, P: pressure, k: permeability, $\phi$: porosity, RF: recovery factor.

### 3.3. Gas RF estimations

The RMSE, CD, and r values for different combinations of the databases used to estimate the gas RF were calculated and presented in Table 6. Comparing the RMSE, CD, and r values reported for the three ML algorithms indicates that, in general, the XGBoost model estimated the gas RF more accurately than the SVM and MLR models for the train datasets, similar to the oil results. For the XGBoost, we found that as the train dataset became larger, the RMSE decreased from 0.190 for the smallest database (i.e., GC) to 0.054 for the largest database (i.e., GCA). For the SVM and MLR, however, the RMSE value for the GCA database was greater than those for the GA and CA databases.

For the test datasets, we found that the accuracy of both XGBoost and SVM models were similar. Although the SVM model estimated the gas RF more accurately than the XGBoost for the GC and GA databases, the XGBoost outperformed the SVM model for the GCA database. Both models provided similar results for the CA database. We also found that the MLR model estimations were not as accurate as the XGBoost and SVM predictions.

Similar to the results of oil RF, the performance of all the algorithms for the independent databases was not satisfactory. As listed in Table 6, the RMSE values for the independent databases are substantially greater than those for the train and test datasets. Recall that the database GCA was the combination of the GasIS, Commercial, and Atlas databases, and, thus, there was no independent database for further evaluations. Based on the RMSE values reported in Table 6, the XGBoost model generally estimated the gas RF for the independent databases more accurately than the SVM and MLR models. Only for the GC database, the MLR model had a slightly better performance than the XGBoost and SVM models. In the assessment of the ML models using the independent databases, we also found CD < 0, similar to the oil RF results (Table 4), which indicate unsatisfactory estimations, confirmed by small r values. The negative CD values obtained for the independent databases (Table 6) suggest that the constructed models did not explain any residual. Instead, such models introduced significantly large additional residuals. In other words, the estimations of the models for the independent databases are worse than raw guessing.

The substantially smaller RMSE, CD, and r values for the independent databases compared to those for the train and test datasets demonstrate the database dependence of the ML performances. We address this in further detail in the Discussion section.

Table 6. Calculated RMSE, CD and r values for the three algorithms XGBoost, SVM, and MLR used to estimate the gas RF. The train datasets of GC, GA, CA, and GCA databases consisted of 5838, 10040, 10251, and 15878 number of samples, respectively.

| Algorithm | XGBoost Regressor ||||||||| 
|---|---|---|---|---|---|---|---|---|---|
| Dataset | Train ||| Test ||| Independent |||
| Database | RMSE | CD | r | RMSE | CD | r | RMSE | CD | r |
| GC | 0.190 | 0.36 | 0.62 | 0.217 | 0.18 | 0.43 | 0.331 | -1.30 | 0.20 |
| GA | 0.057 | 0.93 | 0.97 | 0.093 | 0.82 | 0.90 | 0.694 | -18.70 | -0.1 |
| CA | 0.064 | 0.92 | 0.96 | 0.094 | 0.82 | 0.91 | 0.553 | -5.60 | -0.05 |
| GCA | 0.054 | 0.95 | 0.98 | 0.158 | 0.58 | 0.76 | - | - | - |
| Algorithm | Support Vector Machine Regressor |||||||||
| Dataset | Train ||| Test ||| Independent |||
| Database | RMSE | CD | r | RMSE | CD | r | RMSE | CD | r |
| GC | 0.221 | 0.19 | 0.44 | 0.212 | 0.14 | 0.38 | 0.356 | -1.70 | 0.12 |
| GA | 0.079 | 0.87 | 0.93 | 0.079 | 0.86 | 0.93 | 0.825 | -26.90 | 0.03 |
| CA | 0.083 | 0.86 | 0.93 | 0.094 | 0.82 | 0.91 | 1.342 | -37.70 | 0.12 |
| GCA | 0.160 | 0.56 | 0.75 | 0.165 | 0.54 | 0.74 | - | - | - |
| Algorithm | Stepwise Multiple Linear Regression |||||||||
| Dataset | Train ||| Test ||| Independent |||
| Database | RMSE | CD | r | RMSE | CD | r | RMSE | CD | r |
| GC | 0.220 | 0.14 | 0.37 | 0.228 | 0.10 | 0.32 | 0.316 | -1.10 | 0.34 |

| GA | 0.128 | 0.66 | 0.81 | 0.130 | 0.64 | 0.80 | 1.070 | -45.90 | -0.02 |
| CA | 0.137 | 0.61 | 0.78 | 0.132 | 0.64 | 0.80 | 0.862 | -14.97 | -0.15 |
| GCA | 0.191 | 0.38 | 0.61 | 0.184 | 0.43 | 0.66 | - | - | - |

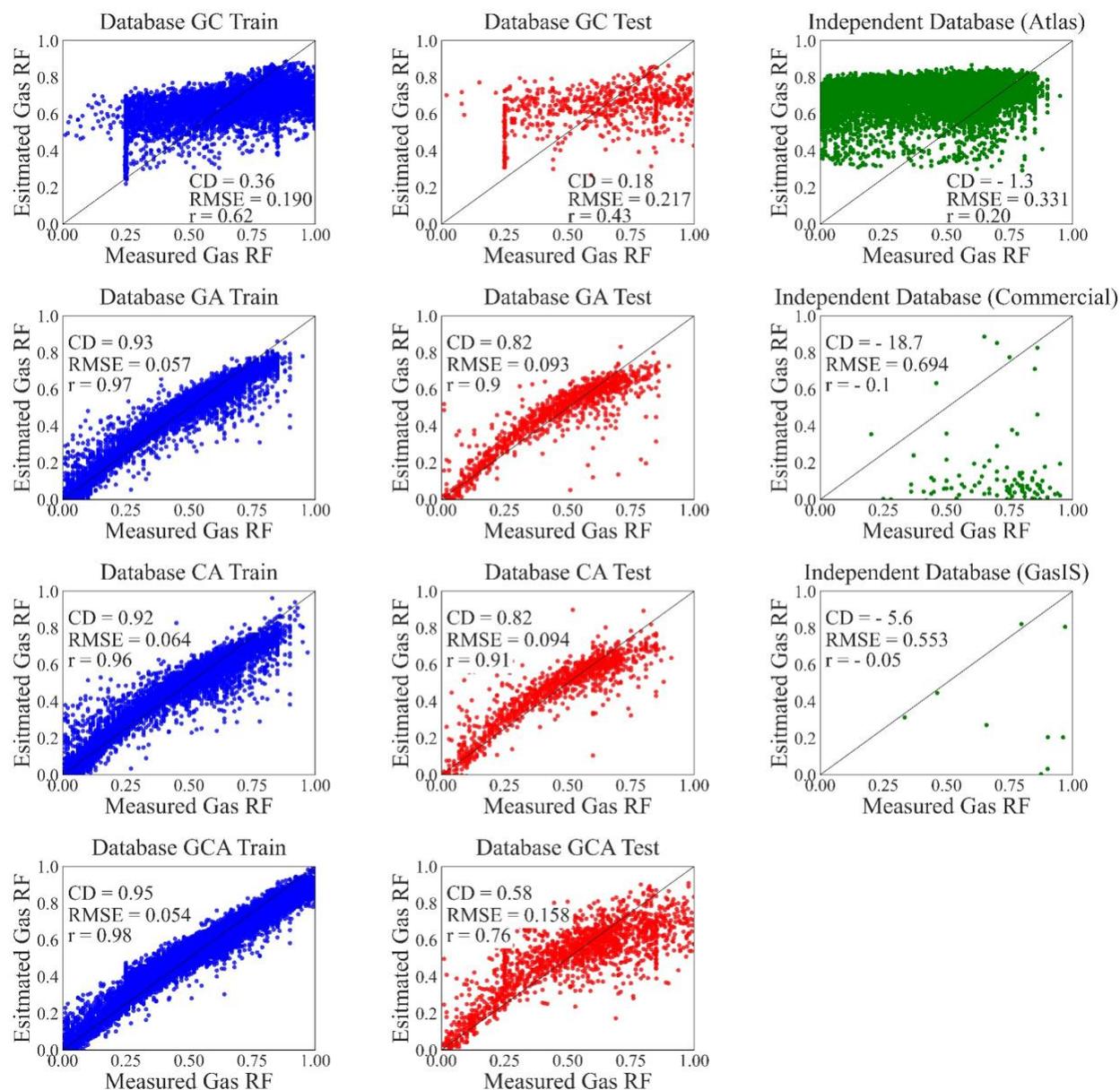

Figure 7. The estimated oil RF by the XGBoost algorithm versus that measured for the train and test datasets as well as the independent database. GC denotes GasIS combined with Commercial, GA is GasIS combined with Atlas, CA represents Commercial combined with Atlas, and GCA is GasIS combined with Commercial and Atlas.

As shown in Figure 7, we found that the constructed XGBoost models tend to overestimate the gas RF at smaller RF values, while underestimate at higher RF values regardless of the data combination set. As stated earlier, this could be explained by the fact that some samples in the train dataset had similar input features, while their gas RFs were different or vice versa. Table 7 lists two samples with similar input features but different gas RF values as well as two other samples with different input features but similar gas RFs.

Table 7. Two samples with similar features but different gas RF (top two) and two samples with different features but similar gas RF (bottom two). Note that the reported values for the input features are not normalized (not scaled between 0 and 1) here.

| GOR (MSCF/RB) | $S_w$ (-) | T (°F) | P (psi) | Thickness (ft) | Reserves (MMSCF) | k (mD) | $\phi$ (ft$^3$/ft$^3$) | Area (acres) | Gas RF |
|---|---|---|---|---|---|---|---|---|---|
| 0 | 0.10 | 160 | 3486 | 75 | $1\times10^6$ | 2 | 0.08 | 25946 | 0.74 |
| 0 | 0.20 | 223 | 4394 | 69 | $7.1\times10^5$ | 2 | 0.05 | 25946 | 0.86 |
| 0 | 0.30 | 230 | 5318 | 1299 | $1.41\times10^9$ | 70 | 0.15 | 1150027 | 0.78 |
| 0 | 0.08 | 268 | 1967 | 299 | $2.12\times10^6$ | 200 | 0.23 | 11120 | 0.74 |

GOR: gas oil ratio, $S_w$: water saturation, T: temperature, P: pressure, k: permeability, $\phi$: porosity, RF: recovery factor.

### 3.4. Feature importance

To investigate the significance of input variables, we performed the feature importance analysis. For the XGBoost, we used SHAP that determines which features have more impact on the constructed model and how the magnitude of features affect model predictions [53]. Figure 8 shows the results of SHAP for the oil and gas RF and the CA (Commercial combined with Atlas) database. Interestingly, the reservoir dimensions (i.e., area and thickness) were detected as two important features in the oil and gas RF estimations. Our results are consistent with those reported by Ghanbarian et al. [54,55] who investigated the effect of scale on capillary pressure curve and permeability estimations via ML. They demonstrated that including sample dimensions (e.g., diameter and length) substantially improved ML estimations at the core scale.

As shown in Figure 8, the constructed models used influential features in petroleum industry (e.g., porosity, $B_o$, and GOR) to estimate the RF. They also shed light on features traditionally not considered to be influential, such as permeability and temperature. Comparing the oil and gas RF graphs shown in Figure 8 implies the similarities in the feature importance selection, which further demonstrates how the pattern recognition in the oil and gas models were similar indicating both models used similar protocols to select important features. This means the workflow used to construct the ML models is logical with no data leakage [26]. Similar feature importance results were obtained for other combinations of databases. For the oil RF and the XGBoost algorithm, the top 4 features selected in the databases TA, CA, and TCA were similar. The feature importance, however, was different in the TC database, which could be due to its limited number of samples. For the gas RF, the top 3 features (i.e., reserves, area, and thickness) were similar in the databases GA, CA, and TCA. Similar to the oil RF results, the GC database had a different feature importance selection (results not shown). This may be because of the small size of this database compared to the other ones.

Results of SHAP showed that the SVM prioritized features very similar to the XGBoost. The top 4 features for oil (i.e., reserves, area, thickness, and GOR) and the top 3 features for gas (i.e., reserves, area, and thickness) were exactly the same variables selected by SHAP and the XGBoost model. This further supports the workflow proposed to construct the ML models.

For the MLR, we investigated the feature importance based on the coefficient of input features, the greater the coefficient the higher the impact on the model. Results showed that the variable reserves was the top feature selected by the MLR algorithm for both oil and gas RFs models except for the TC and GC databases. We found that each MLR model had a different order of feature importance. However, temperature and GOR for gas and temperature and API oil gravity for oil were the commonly-used influential features.

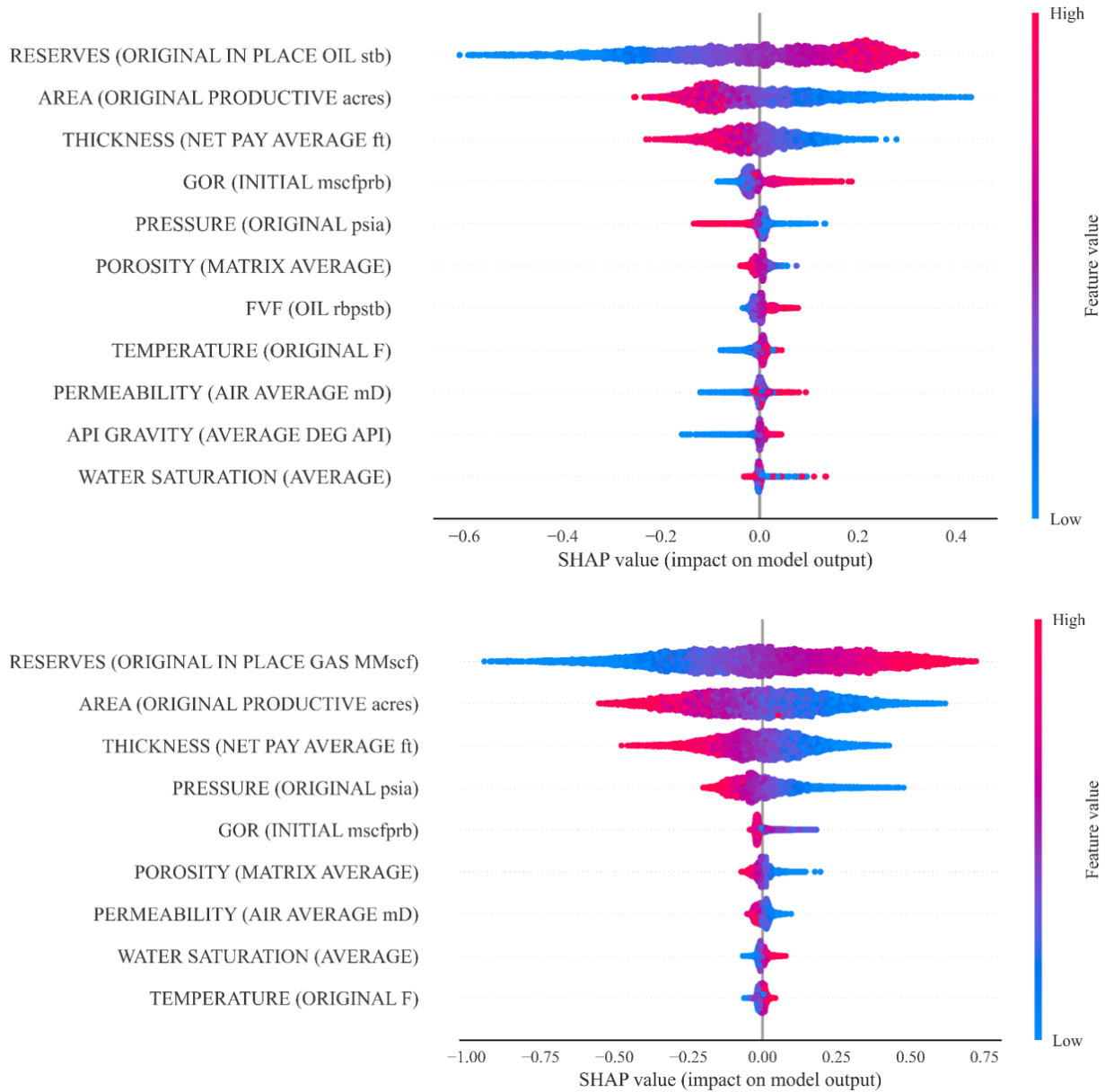

Figure 8. Feature importance of the XGBoost model in the estimation of oil RF (top) and gas RF (bottom) models for the databases CA (with oil RF as target variable) and CA (with gas RF as target variable). Since the data were scaled between 0 and 1, the blue color denotes the smaller values close to 0, while the red color represents greater values near 1. The positive SHAP values indicate positive influence on the model construction and more accurate prediction. However, the negative SHAP values show the negative impact on the model.

## 4. Discussion

As stated earlier, the RF is a function of reservoir's displacement (also known as reservoir drive mechanism or natural drive). Reservoir drive mechanisms are natural forces within a reservoir that move hydrocarbons through it [56]. In the order of efficiency, reservoir drive mechanisms are water drive (bottom or edge), gas cap, solution gas, gravity drainage, and rock or compaction drive. A reservoir may have one or a combination of several drive mechanisms. Drive mechanisms are themselves functions of reservoir qualities, which are capacities of a reservoir to store and transmit oil or gas [57]. The quality of a reservoir is controlled by pore throat size distribution and pore geometry (including natural fractures), pore volume, permeabilities to hydrocarbon, water saturation (hydrocarbon pore volume), lateral continuity, number, position of flow units and containers, reservoir pressure, and drive mechanism [2]. Accordingly, an efficient and reliable ML model should identify reservoir drive mechanism attributes (e.g., pressure and temperature) as well as reservoir quality factors (e.g., porosity and permeability) as the most important features (as shown in Figure 8) to accurately estimate the RF. We should note that the reservoir drive mechanisms were not available for all the databases in this study. However, features attributing to the drive mechanisms and reservoir quality e.g., porosity, thickness, permeability, and water saturation were available.

ML models are developed mainly for prediction purposes. Therefore, they are expected to provide reasonable estimations, if evaluated using a new database including data not used to train and test such models. In the literature, most ML models were not assessed via an independent database. [58] are among the pioneers who addressed the database dependence of ML models and their results. Those authors applied neural networks to train and test models for estimating capillary pressure curve and permeability in soils. Schaap and Leij [58] found that their ML models were optimal for databases based on which models were trained. However, the accuracy and reliability of ML models dropped down when evaluated using other independent databases. Schaap and Leij [58] concluded that the performance of ML models may substantially depend on data used to train and evaluate them. The train-test split is a process through which one would expose a model to unseen data and evaluate the performance of the model. The problem with that is train and test data are still derived from the same database. No matter how randomly a database is split, train and test data would still be similar to some extent.

Another reason for the database-dependent accuracy of ML models is the imperfectness of input variables. ML models strive to compensate for the lack of better features, and such compensation can be reservoir-specific. Therefore, increasing the number of samples collected from different reservoirs in a database may not necessarily improve the performance of ML models.

ML models produce reliable results, if and only if, test and train data are similar both experimentally (e.g., measurement methods and timing of the measurements) and statistically (feature distributions). One should expect the reliability of estimations for test data to drop significantly, if important features were measured at different stages of exploration or even production rather than one single stage. For instance, the feature pressure may vary over time at different stages of a field life. Accordingly, one should consistently use feature values measured at similar life stages of a field using similar methods across all reservoirs. Otherwise, such differences would impact the accuracy of an ML model. For example, for the databases used in this study, it is not clear whether porosity was measured on core samples or determined using sonic or density logs across all the databases. Such differences in measurements may substantially affect the ML results.

As can be deduced from Tables 4 and 6, the performance of the constructed ML models, were unsatisfactory for the independent datasets. Using the t-test method, we statistically analyzed the similarity in the data. More specifically, we compared the features in the train datasets with those in the test datasets as well as those in the independent databases. Results presented for the oil and gas RFs in Table 8 showed no significant differences between the train and test datasets at the 0.05 significance level (p-value > 0.05). However, we found p-values $\ll$ 0.05 when the oil and gas RFs in the train datasets were compared to those in the independent databases (Table 8). This means although the train and test datasets statistically showed similar RF values, the train datasets had significantly different RFs than the independent databases for both oil and gas. Similar results (not shown) were obtained for the input features.

Table 8. p-values determined using the t-test by comparing the RF values in the train datasets with those in the test datasets as well as those in the independent databases for both oil and gas at the significance level of 0.05 (confidence level of 0.95). The *p-values* were calculated using the paired t-test and *unequal variances*.

|  | Database | p–value train and test RFs | Similar data? | p–value train and independent RFs | Similar data? |
|---|---|---|---|---|---|
| Oil | TC | 0.18 | Yes | $2.65 \times 10^{-33}$ | No |
|  | TA | 0.31 | Yes | $2.82 \times 10^{-13}$ | No |
|  | CA | 0.42 | Yes | $1.18 \times 10^{-12}$ | No |
|  | TCA | 0.12 | Yes | - | - |
| Gas | GC | 0.27 | Yes | 0 | No |
|  | GA | 0.51 | Yes | $1.00 \times 10^{-45}$ | No |
|  | CA | 0.65 | Yes | $3.95 \times 10^{-3}$ | No |
|  | GCA | 0.87 | Yes | - | - |

The similarities in the train and test data (Table 8) might be because they come from the same combined database. However, the data in the independent databases are not necessarily similar to those in the train datasets. The dissimilarities in the train and independent data (Table 8) are probably the main reason why performances of the ML models evaluated using the independent datasets were unsatisfactory.

In addition to similarity in the data, we also investigated the similarity in the distributions of top three important input features (i.e., reserves, area, and thickness) as well as the RF. More specifically, using the two-sample Kolmogorov-Smirnov test we determined whether two samples are from the same distribution at the significance level of 0.05. Table 9 presents the results for both oil and gas and the CA database. The statistic values less than the critical value (D) and p-values greater than the significance level indicate that features in the train and test datasets come from the same distribution for both oil and gas. However, we found the statistic values greater than the critical value (D) and p-values substantially less than the significance level for the features in the train datasets and independent databases meaning that they should be from different distributions. Similar results (not shown) were obtained for other database combinations.

Table 9. Results of the Kolmogorov-Smirnov test for the top three input features and RFs in the CA database and the corresponding independent databases TORIS (for oil) and GasIS (for gas). The significance level is 0.05.

| Database | Feature | D train and test datasets | Statistic train and test datasets | p-value train and test datasets | Samples from same distributions ? | D train dataset and independent database | Statistic train dataset and independent database | p-value train dataset and independent database | Samples from same distributions ? |
|---|---|---|---|---|---|---|---|---|---|
| CA (oil RF) | RF | 0.063 | 0.001 | 1.00 | Yes | 0.060 | 0.90 | 0.0 | No |
|  | Reserves |  | 0.024 | 0.95 | Yes |  | 0.81 | 0.0 | No |
|  | Area |  | 0.027 | 0.88 | Yes |  | 0.47 | $4.59 \times 10^{-104}$ | No |
|  | Thickness |  | 0.039 | 0.48 | Yes |  | 0.32 | $2.04 \times 10^{-46}$ | No |
| CA (gas RF) | RF | 0.042 | 0.001 | 1.00 | Yes | 0.017 | 0.92 | $3.33 \times 10^{-10}$ | No |
|  | Reserves |  | 0.028 | 0.40 | Yes |  | 0.55 | 0.0 | No |
|  | Area |  | 0.016 | 0.96 | Yes |  | 0.38 | 0.0 | No |
|  | Thickness |  | 0.035 | 0.15 | Yes |  | 0.25 | 0.0 | No |

D: Critical Value for the Two-sample Kolmogorov-Smirnov test (2-sided)

We recommend that before one uses the constructed ML models to estimate the oil or gas RF, statistical tests should be conducted to investigate whether the distributions of input features in our train datasets are statistically similar to those in a new database for which the oil or gas RF estimations are desired. If the RF estimation is required for one sample (or reservoir), then the values of important features should be close to the values of means in the train dataset studied here.

## 5. Conclusion

In this study, we used a total of 4 databases (Commercial, TORIS, Atlas, and GasIS) to construct machine learning models for the estimation of oil and gas recovery factors at the reservoir scale. For this purpose, we applied different combinations of the databases and three algorithms i.e., XGBoost, SVM, and MLR. Generally speaking, the XGBoost model estimated both oil and gas RFs more accurately than the SVM and MLR models for the train and test datasets. However, the performances of the three models were unsatisfactory when evaluated using the independent databases. Results showed that the constructed ML models and their performances depend on the databases used to train them. The statistical tests disclosed that the unsatisfactory accuracies of the ML models were because the input and output features in the train datasets were significantly different from those in the independent databases. We recommend evaluating similarities in the input features and their distributions before applying the constructed ML models to estimate the oil or gas RF.

**Acknowledgement**

The authors are grateful to Frank Male, University of Texas at Austin, and Amirhossein Yazdavar, United Healthcare INC, for fruitful discussions and comments. Alireza Roustazadeh acknowledges

Department of Geology, Kansas State University, for financial supports through William J. Barret Fund for Excellence in Geology and Gary and Kathie Sandlin Geology Scholarship. Behzad Ghanbarian acknowledges Kansas State University for support through faculty start-up fund. The Python codes as well as means, standard deviations, and distributions of input features used in this study are available at: https://github.com/alirezaro93/Estimating-hydrocarbon-recovery-factor-at-reservoir-scale-via-machine-learning